\documentclass[letterpaper, 10 pt, conference]{ieeeconf}  

\IEEEoverridecommandlockouts                              

\input{paper_style.sty}

\title{\LARGE \bf
Clarke Coordinates Are Generalized Improved State Parametrization\\for Continuum Robots
}

\author{Reinhard M.~Grassmann and Jessica Burgner-Kahrs
\thanks{We acknowledge the support of the Natural Sciences and Engineering Research Council of Canada (NSERC), [RGPIN-2019-04846].}
\thanks{All authors are with Continuum Robotics Laboratory, Department of Mathematical and Computational Sciences, University of Toronto, Mississauga, ON L5L 1C6, Canada {\tt\small reinhard.grassmann@utoronto.ca}}%
}

\begin{document}

\maketitle
\thispagestyle{empty}
\pagestyle{empty}

\begin{abstract}

In this letter, we demonstrate that previously proposed improved state parameterizations for soft and continuum robots are specific cases of Clarke coordinates. 
By explicitly deriving these improved parameterizations from a generalized Clarke transformation matrix, we unify various approaches into one comprehensive mathematical framework. 
This unified representation provides clarity regarding their relationships and generalizes them beyond existing constraints, including arbitrary joint numbers, joint distributions, and underlying modeling assumptions. 
This unification consolidates prior insights and establishes Clarke coordinates as a foundational tool, enabling systematic knowledge transfer across different subfields within soft and continuum robotics.

\end{abstract}


\section{Introduction}

Selecting an appropriate state parameterization is crucial in the modeling and control of continuum robots, as it should capture the geometric characteristics of the robot while providing clear advantages over alternative parameterizations. 
Consequently, multiple improved parameterizations have been proposed for various continuum robot morphologies. 
For single-segment continuum robots, irrespective of their specific actuation mechanisms or the number of actuators, the workspace forms a two-degree-of-freedom (\SI{2}{dof}) dome in task space. 
Thus, improved state parameterizations commonly employ two arc parameters, resulting in a \SI{2}{dof} representation.

Allen~\textit{et al.} \cite{AllenAlbert_et_al_RoboSoft_2020} exploit the geometric observation that the transformation from the proximal to distal end of a constant-curvature arc can be represented as a pure rotation, identifiable through at least three length measurements — a result similarly observed by Simaan~\textit{et al.} \cite{SimaanTaylorFlint_ICRA_2004}. 
Allen~\textit{et al.} \cite{AllenAlbert_et_al_RoboSoft_2020} show that their parameterizations for $n = 3$ and $n = 4$ facilitate closed-form, computationally efficient forward kinematics and Jacobian, eliminate coordinate singularities in path planning, and yield more intuitive torque computations.

Della Santina~\textit{et al.} \cite{DellaSantinaBicchiRus_RAL_2020} and Dian~\textit{et al.} \cite{DianGuo_et_al_Access_2022} exploit well-known solutions of the robot-dependent mapping for $n = 3$ and $n = 4$ joints, relying on the widely used constant curvature assumption \cite{WebsterJones_IJRR_2010}. 
Both employ non-linear combinations of the arc parameters - an approach also advocated by Dupont~\textit{et al.} \cite{DupontRucker_et_al_JPROC_2022}, rather than the standard arc parameterization popularized by Webster \& Jones~\cite{WebsterJones_IJRR_2010}.
For a pneumatic-actuated soft robot, Della Santina~\textit{et al.} \cite{DellaSantinaBicchiRus_RAL_2020} show that their improved parametrization enhances model-based control due to well-defined forward kinematics, Jacobian, and inertia matrix even in straight configuration.
Dian \textit{et al.} \cite{DianGuo_et_al_Access_2022} validate analogous benefits for cable-driven continuum robots.

Expanding beyond conventional geometric arguments, Grassmann~\textit{et al.} \cite{GrassmannSenykBurgner-Kahrs_arXiv_2024} recently exploited an analogy between constraints inherent in certain continuum robots and Kirchhoff's current law from elertical circuits.
Through this analogy, they derive a generalized Clarke transformation matrix and introduce Clarke coordinates as novel improved state parameterizations.
Although a similar generalized Clarke transformation was previously suggested by Janaszek \cite{Janaszek_PIE_2016}, its applicability to continuum robot modeling had not been previously explored. 
Grassmann~\textit{et al.} \cite{GrassmannSenykBurgner-Kahrs_arXiv_2024} demonstrate that Clarke coordinates provide linear, compact, and computationally efficient solutions for kinematics, sampling, and control, notably without strictly requiring constant curvature assumptions (except in kinematic analyses).
Furthermore, the Clarke coordinates are applicable for $n \geq 3$ joints.

In this letter, we demonstrate that Clarke coordinates unify and generalize all previously proposed improved state parametrizations, including those by Allen \textit{et al.} \cite{AllenAlbert_et_al_RoboSoft_2020}, Della Santina \textit{et al.} \cite{DellaSantinaBicchiRus_RAL_2020}, and Dian \textit{et al.} \cite{DianGuo_et_al_Access_2022}.
This is achieved through the derivation of a linear relationship explicitly linking these parameterizations to the generalized Clarke transformation matrix.
Additionally, we clarify several misconceptions associated with improved state parameterizations, particularly regarding redundant joints and actuation constraints. 
In particular, the following contributions are made:
\begin{itemize}
    \item Derivation of existing improved state parameterizations from the Clarke transformation.
    \item Clarification that none of these parametrizations inherently require the constant curvature assumption.
    \item Identification and clarification of limitations when normalized displacement-actuated joint values (\textit{e.g.}, constant curvature kinematics) are enforced.
\end{itemize}

\section{Clarke Transform}

\begin{figure}
    \vspace{.75em}
    \centering
    \includegraphics[width=0.8\columnwidth]{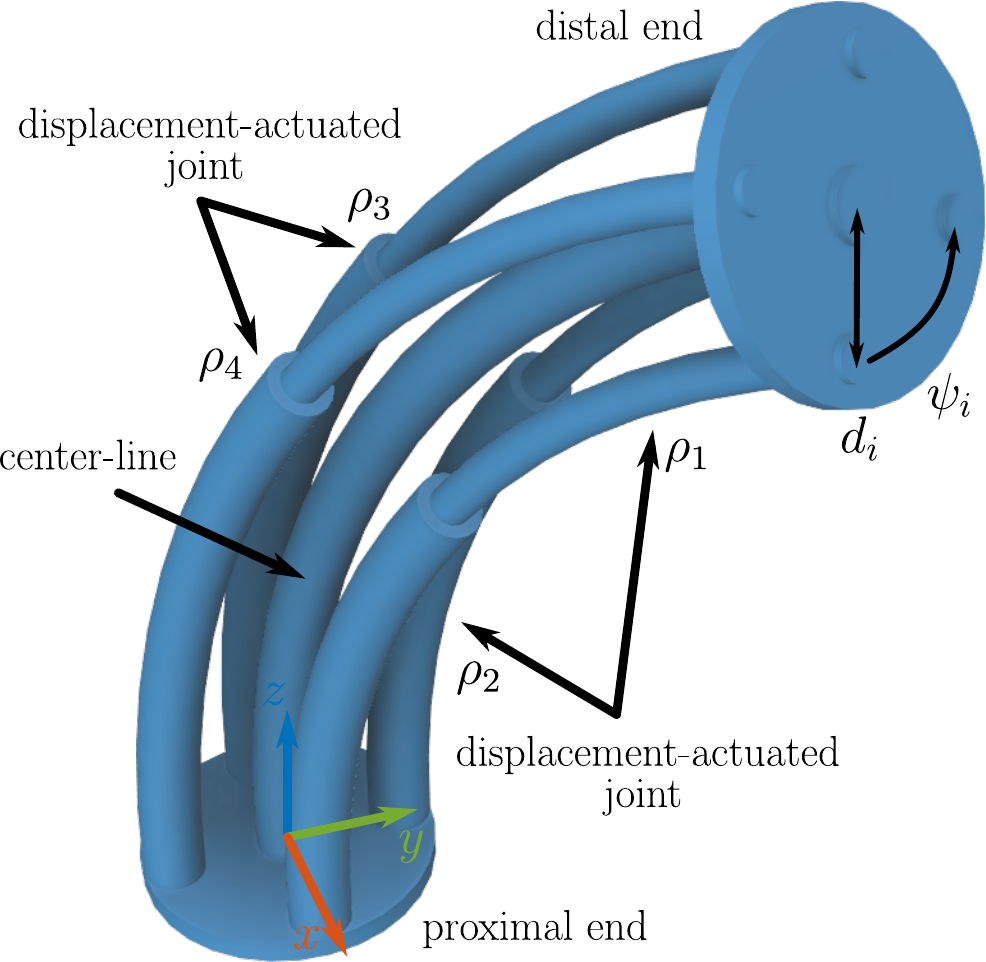}
    \caption{
        Displacement-actuated continuum robot.
        It has $n = 4$ joints, where $i\textsuperscript{th}$ joint location is defined by $\psi_i = 2\pi\left(i - 1\right)/n$ and $d_i = d$.
    }
    \label{fig:dacr}
\end{figure}

In this section, we provide a self-contained description of the Clarke transform.
First, we describe the considered robot type introduced in \cite{GrassmannSenykBurgner-Kahrs_arXiv_2024}.
It is depicted in Fig.~\ref{fig:dacr}.
Second, we summarize the intuition and analogy provided in \cite{GrassmannSenykBurgner-Kahrs_arXiv_2024}.
Afterward, possible generalized Clarke transformation matrices are briefly discussed.
The section ends with a brief recap on the Clarke transform \cite{GrassmannSenykBurgner-Kahrs_arXiv_2024} and our take on the manifold.

\subsection{Displacement-Actuated Continuum Robot}

The abstraction called displacement-actuated continuum robot is introduced by Grassmann \textit{et al.} \cite{GrassmannSenykBurgner-Kahrs_arXiv_2024} to unify the kinematic description of a wide set of soft and continuum robots.
The kinematic design parameters fully describe it.
The segment length $l$ is constant.
Assuming fully constrained actuation paths, the distance $d_i$ and angle $\psi_i$ of each joint location on the cross-section is constant along arc length.
Furthermore, the $n$ joints are equally distributed. 
The only variables are the joint values described by displacements $\rho_i$.

All joint values $\rho_i$ are grouped in an index notation, \textit{i.e.},
\begin{align}
    \rhovec = \left( \rho_i \right)_i = \left( \rho_\text{Re}\cos{\psi_i} + \rho_\text{Im}\sin{\psi_i} \right)_i \subset \mathbb{R}^n
    \label{eq:rho}
\end{align}
as the vector $\rhovec$, where the free parameters are combined into
\begin{align}
    \rhoclarke = \left[ \rhoreal, \rhoim \right]\transpose \in \mathbb{R}^2.
    \label{eq:rho_clarke}
\end{align}
This indicates that the displacement-actuated continuum robot can be described with two variables using \eqref{eq:rho_clarke} instead of $n$ in \eqref{eq:rho}.
In fact, the displacement constraint
\begin{align}
	\sum_{i=1}^{n} \rho_{i} = 0, 
    \label{eq:sum_rho}
\end{align}
indicates that all joint values are interdependent.
Interestingly, \eqref{eq:sum_rho} is often not mentioned, \textit{e.g.}, \cite{AllenAlbert_et_al_RoboSoft_2020, DellaSantinaBicchiRus_RAL_2020, DianGuo_et_al_Access_2022}, or unfortunately stated using actuation length, \textit{e.g.}, \cite{WebsterJones_IJRR_2010, RaoBurgner-Kahrs_et_al_Frontiers_2021}.

\subsection{Intuition Using Generic Clarke Transformation Matrix}
\label{sec:intuition}

The Clarke transform utilized a generalization of a Clarke transformation matrix $\boldsymbol{M}_\text{Clarke}$.
A generic form as ${3 \times 3}$ matrix is given by
\begin{align}
	\boldsymbol{M}_\text{Clarke}
    = k_0
    \begin{bmatrix} 
        1 & -\dfrac{1}{2} & -\dfrac{1}{2} \\[1em]
        0 & \dfrac{\sqrt{3}}{2} & -\dfrac{\sqrt{3}}{2} \\[1em]
        k_1 & k_1 & k_1 \\
    \end{bmatrix}
    \in \mathbb{R}^{3 \times 3}
    ,
    \label{eq:M_Clarke}
\end{align}
where $k_0$ and $k_1$ are free parameters.
In electrical engineering, \eqref{eq:M_Clarke} with the right parameters for $k_0$ and $k_1$ can used to project three-phase quantities, \textit{e.g.}, the electric currents $i_\mathrm{a}$,$i_\mathrm{b}$, and $i_\mathrm{c}$, onto the plane defined by $i_\mathrm{a} + i_\mathrm{b} + i_\mathrm{c} = 0$.
For a balanced system, the plane's coordinates are $i_\alpha$ and $i_\beta$, where a third variable reduces to $i_\gamma = 0$.
The mapping
\begin{align}
    \begin{bmatrix} 
        i_\alpha & i_\beta & 0
    \end{bmatrix}\transpose
    = 
	\boldsymbol{M}_\text{Clarke}
    \begin{bmatrix} 
        i_\mathrm{a} & i_\mathrm{b} & i_\mathrm{c}
    \end{bmatrix}\transpose
    \nonumber
\end{align}
is used to simplify approaches, \textit{e.g.}, control problems leading to vector control also known as field-oriented control. 
First, note that commonly two variances of \eqref{eq:M_Clarke} are considered, \textit{i.e.}, power-invariant ($k_0 = \sqrt{2/3}$ and $k_1 = \sqrt{2}/2$) and amplitude-invariant forms ($k_0 = 2/3$ and $k_1 = 1/2$).
Second, note that the inverse of \eqref{eq:M_Clarke} allows the recovery of the three-phase quantities from $i_\alpha$ and $i_\beta$ without information loss.
Third, note that \eqref{eq:M_Clarke} can be reduced to ${2 \times 3}$ matrix, if $i_\gamma = 0$ holds.
Finally, note that the three-phase quantities are related to current or voltage.
Using a more general description, \eqref{eq:M_Clarke} can be used for flow or effort quantities, where flow and effort relate to current and voltage, respectively. 

The intuition continues by recognizing that Kirchhoff's current law for three phases, \textit{i.e.}, $i_\mathrm{a} + i_\mathrm{b} + i_\mathrm{c} = 0$, is a stark reminder of the displacement constraint \eqref{eq:sum_rho} for three displacements.
Loosely speaking, a phase, \textit{e.g.}, $i_\mathrm{a}$, represents a displacement, \textit{e.g.}, $\rho_1$.
Furthermore, $i_\alpha$ and $i_\beta$ relates to $\rhoreal$ and $\rhoim$, respectively.
We kindly refer to Grassmann \textit{et al.} \cite{GrassmannSenykBurgner-Kahrs_arXiv_2024} to further elaboration on the intuition and analogy.
To conclude, utilizing and generalizing \eqref{eq:M_Clarke} appears promising.

\subsection{Generalized Clarke transformation and its Inverse}

To consider $n$-phases, different generalizations of \eqref{eq:M_Clarke} exist. 
Squared $n \times n$ matrices are proposed by Willems \cite{Willems_TOE_1969} and by Rockhill \& Lipo \cite{RockhillLipo_IEMDC_2015}.
However, in the ideal case, $\left(n - 2\right)$ zeros are computed and, therefore, a squared $n \times n$ matrix should be reducible to a $n \times 2$ matrix or $2 \times n$ matrix.
Note that some authors, \textit{e.g.}, \cite{Willems_TOE_1969, EboulePretoriusMbuli_APPEEC_2019}, define \eqref{eq:M_Clarke} as the inverse transformation.
Janaszek \cite{Janaszek_PIE_2016} and Grassmann \textit{et al.} \cite{GrassmannSenykBurgner-Kahrs_arXiv_2024} provide similar $n \times 2$ matrices, where one matrix is the negative of the other matrix.
The latter generalization is preferred as it can be directly related to the arc parameters of the arc space.

Grassmann \textit{et al.} \cite{GrassmannSenykBurgner-Kahrs_arXiv_2024} define the generalized Clarke transformation matrix $\boldsymbol{M}_\mathcal{P} \in \mathbb{R}^{2 \times n}$ for $n$ displacements as
\begin{flalign}
	\MP
    =
    \dfrac{2}{n}\!
	\begin{bmatrix}
		\cos\left(\psi_1\right) & \cos\left(\psi_2\right) & \cdots & \cos\left(\psi_n\right)\\[0.5em]
		\sin\left(\psi_1\right) & \sin\left(\psi_2\right) & \cdots & \sin\left(\psi_n\right)
	\end{bmatrix}
    .
	\label{eq:MP}
\end{flalign}
Its inverse is the right-inverse of  $\boldsymbol{M}_\mathcal{P}$ defined as 
\begin{align}
	\MPinv
    =
	\begin{bmatrix}
		\cos\left(\psi_1\right) & \sin\left(\psi_1\right) \\
        \cos\left(\psi_2\right) & \sin\left(\psi_2\right) \\
        \vdots & \vdots\\
        \cos\left(\psi_n\right) & \sin\left(\psi_n\right)
	\end{bmatrix}
    ,
	\label{eq:MP_inverse}
\end{align}
which can be directly found by recognizing \eqref{eq:rho} as a system of linear equations \cite{GrassmannBurgner-Kahrs_arXiv_2024a}.
The notation in \eqref{eq:MP_inverse} differs from the possibly ambiguous notation used in \cite{GrassmannSenykBurgner-Kahrs_arXiv_2024, GrassmannBurgner-Kahrs_arXiv_2024a, GrassmannBurgner-Kahrs_ICRA_EA_2024, GrassmannBurgner-Kahrs_arXiv_2024b}.
Note the simple relation between \eqref{eq:MP} and \eqref{eq:MP_inverse}, \textit{i.e.}, $\MPinv = \left(n/2\right)\MP\transpose$.
Furthermore, note that $\MP \MPinv = \Imat_{2 \times 2}$ and $\MPinv \MP \not= \Imat_{n \times n}$ with $\det\left(\MPinv \MP\right) = 0$.
The superscript $\text{\sffamily R}$ in $\MPinv$ is a notation, not an operator. 
The notation should remind the practitioner that \eqref{eq:MP_inverse} acts like a right inverse matrix when the production of \eqref{eq:MP} and \eqref{eq:MP_inverse} is evaluated.

\subsection{Clarke Transform and the Two-Dimensional Manifold}

Applying \eqref{eq:MP} and \eqref{eq:MP_inverse}, the Clarke coordinates $\rhoclarke$ and displacement $\rhovec \in \jointspace$ can be set in relation to each other, \textit{i.e.},
\begin{align}
    \rhoclarke &= \MP\rhovec
    \quad\text{and}\quad
    \label{eq:forward}
    \\
    \rhovec &= \MPinv\rhoclarke
    .
    \label{eq:inverse}
\end{align}
where $\MPinv$ is the right-inverse of $\MP$ and $\MP$ is given by the generalized Clarke transformation matrix.
According to Grassmann~\textit{et al.} \cite{GrassmannSenykBurgner-Kahrs_arXiv_2024}, Clarke transform is the mapping of quantities using \eqref{eq:forward} and \eqref{eq:inverse}, where former is the forward transform and the latter is the inverse transform.
For a symmetric arranged joint location, we kindly refer to \cite{GrassmannSenykBurgner-Kahrs_arXiv_2024} for a rigor derivation of \eqref{eq:forward} and \eqref{eq:inverse}.
Note that, for non-symmetric arranged joint location, \eqref{eq:forward} is different, whereas \eqref{eq:inverse} remains the same, see Grassmann \& Burgner-Kahrs \cite{GrassmannBurgner-Kahrs_arXiv_2024a} for details.

It is important to highlight that the elements $\rho_i$ of $\rhovec$ are interdependent.
Therefore, $\rhovec$ is not properly defined by $\mathbb{R}^n$ as not all elements of $\mathbb{R}^n$ can be a $\rhovec$.
For clarification,
\begin{align}
    \rhovec \in \jointspace \subset \mathbb{R}^n
    \quad\text{and}\quad
    \rhoclarke \in \mathbb{R}^2
    ,
    \nonumber
\end{align}
where $\jointspace$ is the joint space defining the set of all valid joint values $\rhovec$.
According to Grassmann \textit{et al.} \cite{GrassmannSenykBurgner-Kahrs_arXiv_2024}, it is defined as
\begin{align}
    \jointspace = 
    \left\{ \rhovec \in \mathbb{R}^n \ \bigl| \ \rhovec = \MPinv\rhoclarke \ \wedge \ \rhoclarke \in \mathbb{R}^2\right\},
    \label{eq:jointspace}
\end{align}
where $\MPinv$ is robot-specific and defined in \eqref{eq:MP_inverse}.

In \eqref{eq:jointspace}, we interpret \eqref{eq:inverse} as a set of $n$ constraints. 
Rewriting the constraints to $\mathbf{0}_{n \times 1} = \MPinv\rhoclarke - \rhovec$, we can find $n$ equations each describing a plane, \textit{i.e.}, $0 = \rho_\text{Re}\cos{\psi_i} + \rho_\text{Im}\sin{\psi_i} - \rho_i$, in the uplifted space $\mathbb{R}^{n+2}$.
The intersection of all planes is a plane spanned by the Clarke coordinates.

By inspection of \eqref{eq:rho}, we span $\jointspace$ using two vectors
\begin{align}
    \vv_1 &= 
    \MPinv\hotvec_{2 \times 1}^{(1)}
    = 
	\begin{bmatrix}
		\cos\left(\psi_1\right) &
        \cos\left(\psi_2\right) &
        \cdots &
        \cos\left(\psi_n\right)
	\end{bmatrix}
    \transpose
    \nonumber
    \\
    \text{and}
    \nonumber
    \\
    \vv_2 &= 
    \MPinv\hotvec_{2 \times 1}^{(2)}
    = 
	\begin{bmatrix}
		\sin\left(\psi_1\right) &
        \sin\left(\psi_2\right) &
        \cdots &
        \sin\left(\psi_n\right)
	\end{bmatrix}
    \transpose
    \nonumber
\end{align}
with $\hotvec_{2 \times 1}^{(1)} = \left[	1, 0\right]\transpose$ and $\hotvec_{2 \times 1}^{(2)} = \left[ 0, 1\right]\transpose$.
This leads to
\begin{align}
    \jointspace = 
    \bigl\{\bigr.\,&\rhoreal\vv_1 + \rhoim\vv_2 \,\, \bigl|\bigr. \,\, \rhoreal, \rhoim \in \mathbb{R} \,\wedge\nonumber\\    
    \,& \vv_1 = \MPinv\hotvec_{2 \times 1}^{(1)} \,\wedge\, \vv_2 = \MPinv\hotvec_{2 \times 1}^{(2)} \,\bigl.\bigr\}
    .
    \label{eq:jointspace_alt_w_span}
\end{align}
Our alternative description \eqref{eq:jointspace_alt_w_span} has the benefit that the degrees of freedom of the system can be seen directly as this is the number of independent vectors in $\jointspace$.
Using the trigonometric identity stated in \cite{GrassmannSenykBurgner-Kahrs_arXiv_2024, GrassmannBurgner-Kahrs_ICRA_EA_2024}, we can show that
\begin{align}
	\vv_1\transpose\vv_2 = \sum_{i=1}^{n} \sin\left(\psi_i\right)\cos\left(\psi_i\right) = 0
\end{align}
and, therefore, both vectors are orthogonal and independent.
Note that their squared magnitude is $\vv_1\transpose\vv_1 = \vv_2\transpose\vv_2 = n/2$.

Finally, we present a second alternative to describe $\jointspace$ without relying on the Clarke coordinates. 
We define
\begin{align}
    \jointspace = 
    \left\{ \rhovec \in \mathbb{R}^n \ \bigl| \ \rhovec = \MPinv\MP\rhovec \ \right\}
    ,
    \label{eq:jointspace_alt_wo_ClarkeCoordinates}
\end{align}
where \eqref{eq:forward} and \eqref{eq:inverse} are used to constrain $\rhovec$.

\section{Relation to Improved State Parameterizations}

In this section, we derive the improved state parametrizations proposed by Della Santina \textit{et al.} \cite{DellaSantinaBicchiRus_RAL_2020}, Allen \textit{et al.} \cite{AllenAlbert_et_al_RoboSoft_2020}, and Dian \textit{et al.} \cite{DianGuo_et_al_Access_2022} using Clarke transform.
First, we compare actuation length and displacement actuation.
Afterward, we present the derivations and their relation to the Clarke coordinates.
Table~\ref{tab:overview_improved_state_representation} lists their notation and the found relations.

\begin{table*}[!t]
    \vspace{.75em}
    \renewcommand*{\arraystretch}{1.4}
    \caption{
        Various improved state parametrizations.
        Variable names are taken from the literature and the notations are adapted to our notation. 
    }
    \label{tab:overview_improved_state_representation}
    \centering
    \begin{tabular}{r r rr rr} 
        \toprule
        \multicolumn{1}{N}{Reference} & 
        \multicolumn{1}{N}{$n$} & 
        \multicolumn{2}{N}{Parameterization \textit{w.r.t.} joint values} & 
        \multicolumn{2}{N}{Parameterization \textit{w.r.t.} Clarke Coordinates} \\
		\cmidrule(r){1-1}
		\cmidrule(lr){2-2}
		\cmidrule(lr){3-4}
		\cmidrule(lr){5-6}
        \\
        Della Santina \textit{et al.} \cite{DellaSantinaBicchiRus_RAL_2020} & $4$ & $\Delta_{x} = \dfrac{l_{3} - l_{1}}{2}$ & $\Delta_{y} = \dfrac{l_{4} - l_{2}}{2}$ & $\Delta_{x} = \rhoreal$ & $\Delta_{y} = \rhoim$ \\[.75em]
        Allen \textit{et al.} \cite{AllenAlbert_et_al_RoboSoft_2020} & 3 & $u = \dfrac{l_2 - l_3}{\sqrt{3}d}$ & $v = \dfrac{\left(l_1 + l_2 + l_3\right)/3 - l_1}{d}$ & $v = \left(1/d\right)\rhoreal$ & $u = \left(1/d\right)\rhoim$ \\[.75em]
        Allen \textit{et al.} \cite{AllenAlbert_et_al_RoboSoft_2020} & 4 & $u = \dfrac{l_2 - l_4}{d}$ & $v = \dfrac{l_3 - l_1}{d}$ & $v = \left(2/d\right)\rhoreal$ & $u = \left(2/d\right)\rhoim$ \\[.75em]
        Dian \textit{et al.} \cite{DianGuo_et_al_Access_2022} & 3 & $\Delta x = \dfrac{l_2 + l_3 - 2l_1}{3}$ & $\Delta y = \dfrac{l_3 - l_2}{\sqrt{3}}$ & $\Delta x = \rhoreal$ & $\Delta y = \rhoim$ \\[.75em]
		\cmidrule(r){1-1}
		\cmidrule(lr){2-2}
		\cmidrule(lr){3-4}
		\cmidrule(lr){5-6}
        Grassmann \textit{et al.} \cite{GrassmannSenykBurgner-Kahrs_arXiv_2024} & $n$ & \multicolumn{2}{c}{$\left[ \rhoreal, \rhoim \right]\transpose = \MP\rhovec$} & $\rhoreal$ & $\rhoim$ \\[.25em]
        \bottomrule
    \end{tabular}
\end{table*}

\subsection{Absolute Length or Displacement as Actuation?}

In the literature, \textit{e.g.}, \cite{AllenAlbert_et_al_RoboSoft_2020, DellaSantinaBicchiRus_RAL_2020, DianGuo_et_al_Access_2022}, and in reviews, \textit{e.g.}, \cite{WebsterJones_IJRR_2010, RaoBurgner-Kahrs_et_al_Frontiers_2021}, the absolute length $l_i$ is preferred over displacement $\rho_i$.
Neglecting the constant curvature assumption, 
\begin{align}
    l_i = l - \rho_i
    \label{eq:actuation_literature}
\end{align}
describes the relation between both, where $l$ is the segment length.
Unfortunately, \eqref{eq:actuation_literature} is used in most improved state parametrizations, \textit{e.g.}, \cite{AllenAlbert_et_al_RoboSoft_2020, DellaSantinaBicchiRus_RAL_2020, DianGuo_et_al_Access_2022}, obstructing possible extensions to $n$ joints.
In the following, we will show that carrying a constant offset, \textit{i.e.}, $l$, does not affect any given improved state parametrization. 
Due to $\MP\onevec_{n \times 1} = \mathbf{0}_{2 \times 1}$, where $\onevec_{n \times 1}$ has ones everywhere and $\mathbf{0}_{2 \times 1}$ has zeros everywhere, constant values are filtered out \cite{GrassmannSenykBurgner-Kahrs_arXiv_2024, GrassmannBurgner-Kahrs_ICRA_EA_2024}.
Therefore, $l$ in a vectorized form of \eqref{eq:actuation_literature} maps to the zero vector.

\subsection{Relation to Dian \textit{et al.}}

First, we express $\Delta x$ and $\Delta y$ using $\rhovec$ for $n = 3$ and \eqref{eq:actuation_literature}.
This leads to 
\begin{align}
    \Delta x &= \dfrac{l_2 + l_3 - 2l_1}{3} = \dfrac{- \rho_2 - \rho_3 + 2\rho_1}{3}
    \quad\text{and}
    \nonumber\\
    \Delta y &= \dfrac{l_3 - l_2}{\sqrt{3}} = \dfrac{\rho_2 - \rho_3}{\sqrt{3}}
    ,
    \nonumber
\end{align}
where the constant $l$ is canceled out.
Afterward, using the generic form $\boldsymbol{M}_\text{Clarke}$ stated in \eqref{eq:M_Clarke} with parameter $k_0$ and $k_1$, we set up a set of linear equations leading to
\begin{align}
    \begin{bmatrix} 
        \Delta x \\[1em]
        \Delta y \\[1em]
        0 \\
    \end{bmatrix}
    =
    \begin{bmatrix} 
        k_0\dfrac{2\rho_1 - \rho_2 - \rho_3}{2} \\[0.5em]
        k_0\dfrac{\sqrt{3}\rho_2 - \sqrt{3}\rho_3}{2} \\[0.5em]
        k_0 k_1 \left(\rho_1 + \rho_2 + \rho_3\right) \\
    \end{bmatrix}
    = 
    \boldsymbol{M}_\text{Clarke}\rhovec
    .
    \nonumber
\end{align}
The last row is zero due to \eqref{eq:sum_rho}.
To recover $\Delta x$ and $\Delta y$, we can set $k_0 = 2/3 = 2/n$ and $k_1 \in \mathbb{R}$.
This is exactly
\begin{align}
    \begin{bmatrix} 
        \Delta x \\
        \Delta y \\
    \end{bmatrix}
    =
    \begin{bmatrix} 
        \rhoreal \\
        \rhoim \\
    \end{bmatrix}
    =
    \rhoclarke
    = 
    \MP\rhovec
    \nonumber
\end{align}
with $\MP$ defined in \eqref{eq:MP} for $n = 3$.
Therefore, we can conclude that the improved state parameterizations proposed by Dian \textit{et al.} \cite{DianGuo_et_al_Access_2022} are Clarke coordinates for $n = 3$.

\subsection{Relation to Della Santina \textit{et al.}}

Della Santina \textit{et al.} \cite{DellaSantinaBicchiRus_RAL_2020} consider $n = 4$, where the indexing of the variables is different, \textit{cf.} Fig.~4 in \cite{DellaSantinaBicchiRus_RAL_2020} and Fig.~\ref{fig:dacr} in this work.
Using \eqref{eq:actuation_literature} and our notation, we can write
\begin{align}
    \Delta_x &= \dfrac{l_3 - l_1}{2} = \dfrac{\rho_1 - \rho_3}{2}
    \quad\text{and}
    \nonumber\\
    \Delta_y &= \dfrac{l_4 - l_2}{2} = \dfrac{\rho_2 - \rho_4}{2}
    .
    \nonumber
\end{align}
Considering $\MP$ and $\rhovec$ for $n = 4$, we can immediately see
\begin{align}
    \begin{bmatrix} 
        \Delta_x \\
        \Delta_y \\
    \end{bmatrix}
    =
    \frac{1}{2}
    \begin{bmatrix} 
        \rho_1 - \rho_3 \\
        \rho_2 - \rho_4 \\
    \end{bmatrix}
    =
    \begin{bmatrix} 
        \rhoreal \\
        \rhoim \\
    \end{bmatrix}
    = 
    \frac{1}{2}
	\begin{bmatrix}
		1 & 0 & -1 & 0\\
		0 & 1 & 0 & -1 
	\end{bmatrix}
    \rhovec
    \nonumber
\end{align}
proofing that the improved state parametrization by Della Santina \textit{et al.} \cite{DellaSantinaBicchiRus_RAL_2020} are Clarke coordinates for $n = 4$.

\subsection{Relation to Allen \textit{et al.}}

Allen \textit{et al.} \cite{AllenAlbert_et_al_RoboSoft_2020} propose two sets of improved state parametrization.
Starting with $n = 3$ and using \eqref{eq:actuation_literature}, we can write
\begin{align}
    u &= \dfrac{l_2 - l_3}{\sqrt{3}d} = \dfrac{\rho_3 - \rho_2}{\sqrt{3}d}
    \quad\text{and}
    \nonumber\\
    v &= \dfrac{\left(l_1 + l_2 + l_3\right)/3 - l_1}{d} = \dfrac{2\rho_1 - \rho_2 - \rho_3}{3d}
    .
    \nonumber
\end{align}
We can already observe that $u$ and $v$ are linear to $\Delta y$ and $\Delta x$, respectively.
The factor is $d$.
Therefore, we conclude that the improved state parameterization by Allen \textit{et al.} \cite{AllenAlbert_et_al_RoboSoft_2020} for $n = 3$ is linearly related to the Clarke coordinates, \textit{i.e.}, $\rhoreal = \Delta x = vd$ and $\rhoim = \Delta y = ud$.

Continuing with the second set of improved state parameterization for $n = 4$, we immediately see that 
\begin{align}
    u &= \dfrac{l_2 - l_4}{d} = \dfrac{\rho_4 - \rho_2}{d} = -\dfrac{\rho_2 - \rho_4}{d}
    \quad\text{and}
    \nonumber\\
    v &= \dfrac{l_3 - l_1}{d} = \dfrac{\rho_1 - \rho_3}{d}
    \nonumber
\end{align}
are linear to the improved state parameterization by Della Santina \textit{et al.} \cite{DellaSantinaBicchiRus_RAL_2020}, when $u$ and $v$ are expressed using \eqref{eq:actuation_literature}.
Due to $\rhoreal = \Delta_x = vd/2$ and $\rhoim = \Delta_y = -ud/2$, we conclude that the state parameterization by Allen \textit{et al.} \cite{AllenAlbert_et_al_RoboSoft_2020} for $n = 4$ is linearly related to the Clarke coordinates.
\section{Note On the Constant Curvature Assumption}

Della Santina~\textit{et al.} \cite{DellaSantinaBicchiRus_RAL_2020}, Dian~\textit{et al.} \cite{DianGuo_et_al_Access_2022}, and Allen~\textit{et al.} \cite{AllenAlbert_et_al_RoboSoft_2020} explicitly rely on the constant curvature assumption for their improved state parameterizations. 
Other authors, such as Cao~\textit{et al.} \cite{CaoXie_et_al_JMR_2022}, directly introduce a nonlinear combination of commonly used arc parameters without explicitly referencing the robot-dependent mapping. 
Such nonlinear combinations can be systematically derived using the Clarke transform and the associated Clarke coordinates. 
Grassmann \textit{et al.}\cite{GrassmannSenykBurgner-Kahrs_arXiv_2024} express this relationship explicitly as:
\begin{align}
    \begin{bmatrix}
        dl\kappa\cos\left(\theta\right) \\ 
        dl\kappa\sin\left(\theta\right)
    \end{bmatrix}
    = 
    \rhoclarke
    = 
    \boldsymbol{M}_\mathcal{P}
    \boldsymbol{\rho}
    ,
    \label{eq:rhoclarke_arc_parameters_constant_curvature}
\end{align}
where $\theta$ denotes the bending plane angle, $\phi = l\kappa$ is the bending angle, and the curvature $\kappa$ is constant \textit{w.r.t.} the arc length $s$.
Equation \eqref{eq:rhoclarke_arc_parameters_constant_curvature} explicitly highlights the nonlinear combination of parameters $\kappa$ and $\theta$, inherent to the constant curvature assumption.

\subsection{Generalization to Non-Constant Curvature}

\begin{figure}
    \vspace{.75em}
    \centering
    \includegraphics[width=\columnwidth]{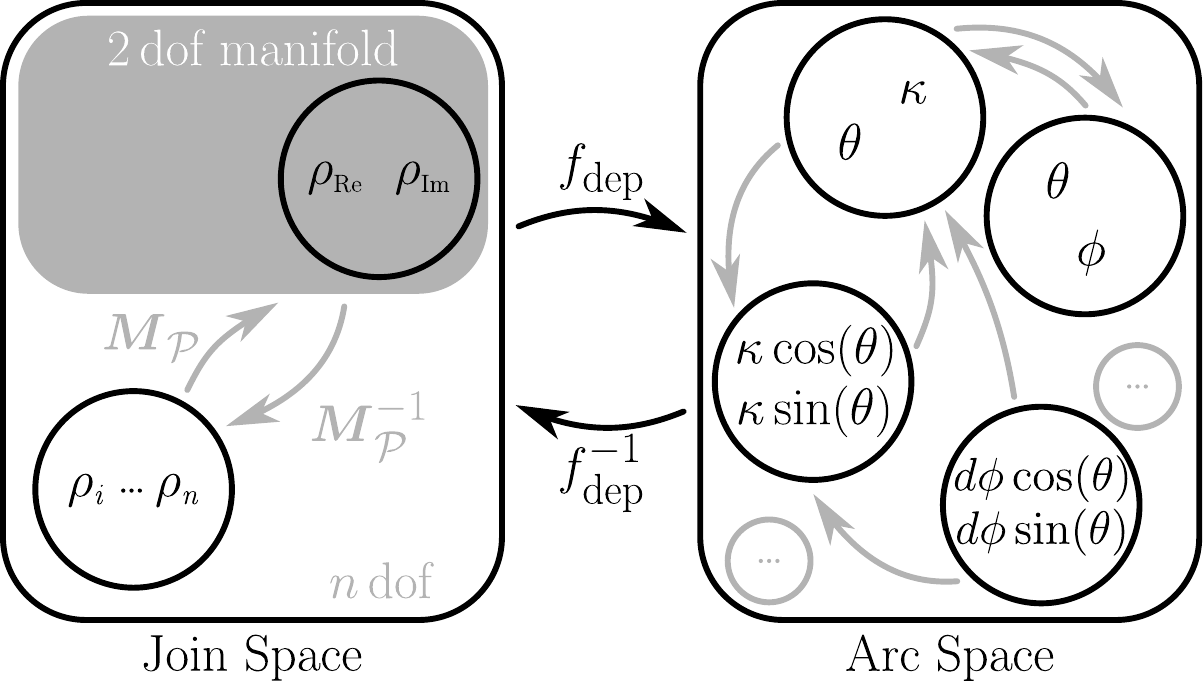}
    \caption{
        Transformation between joint space and arc space.
        Segment length $l$ is omitted since only a non-extensible displacement-actuated continuum robot is considered.
        Therefore, $l$ is one of the kinematic parameters and not an arc parameter.
    }
    \label{fig:joint_space_2_arc_space}
\end{figure}

While the specific formulation in \eqref{eq:rhoclarke_arc_parameters_constant_curvature} assumes constant curvature, neither the general derivation of the Clarke transform nor Clarke coordinates inherently require this assumption. 
Grassmann \textit{et al.}~\cite{GrassmannSenykBurgner-Kahrs_arXiv_2024} explicitly consider only kinematic design parameters and the fundamental constraint given by \eqref{eq:sum_rho}.
In Grassmann \& Burgner-Kahrs~\cite{GrassmannBurgner-Kahrs_arXiv_2024b}, they further demonstrate that this constraint corresponds to the concept of parallel (offset) curves, commonly used in computer graphics.
In planar scenarios, these parallel curves defining the fully constrained actuation paths depend solely on the offset distance $d$ and local curvature $\kappa(s)$.
Leveraging this relationship, Grassmann \& Burgner-Kahrs~\cite{GrassmannBurgner-Kahrs_arXiv_2024b} derive a generalized expression for the tip orientation $\phi$ from actuator displacement $\rhovec$ without imposing constant curvature.
This generalizes earlier results by Simaan~\textit{et al.} \cite{SimaanTaylorFlint_ICRA_2004} and Allen~\textit{et al.} \cite{AllenAlbert_et_al_RoboSoft_2020}, both of which assumed constant curvature. 
We refer the reader to \cite{GrassmannBurgner-Kahrs_arXiv_2024b} for further details.
Therefore, we can formulate 
\begin{align}
    \begin{bmatrix}
        d\phi\cos\left(\theta\right) \\ 
        d\phi\sin\left(\theta\right)
    \end{bmatrix}
    = 
    \rhoclarke
    = 
    \boldsymbol{M}_\mathcal{P}
    \boldsymbol{\rho}
    ,
    \label{eq:rhoclarke_arc_parameters}
\end{align}
that is independent of the constant curvature assumption.
The same parameterization is used by Qu~\textit{et al.} \cite{QuLau_et_al_ROBIO_2016} and Della Santina~\textit{et al.} \cite{DellaSantinaBicchiRus_RAL_2020} for two pairs of differential actuation.
In the following, we want to highlight the benefits of \eqref{eq:rhoclarke_arc_parameters} over other parameterizations of the arc space as shown in Fig.~\ref{fig:joint_space_2_arc_space}.

First, the range of the arc parameters is an important consideration. 
In \eqref{eq:rhoclarke_arc_parameters}, $\theta \in \mathbb{S}^1$ is embedded in $\mathbb{R}^2$ using $\cos\left(\theta\right)$ and $\sin\left(\theta\right)$.
This removes its ambiguity due to $\theta = \theta + 2k\pi$ with $k \in \mathbb{N}$.
To recover $\theta$, one simply uses $\arctantwo\left(\rhoim, \rhoreal\right)$.
Note that $\phi \not\in \mathbb{S}^1$ and, therefore, it is not problematic since $\phi \in \mathbb{R}^+$.
Furthermore, note that many authors assume $\left(\theta, \phi\right) \in \mathbb{R}^2$ which is a misconception on the range of the arc parameters.
Indeed, it is better described by $\left(\theta, \phi\right) \in \mathbb{S}^1\times\mathbb{R}^+$.
Note that $\phi \in \left[0, 2\pi\right]$ might be a suitable restriction as $\phi \leftarrow 2\pi$ would indicate a "full circle" and self-collision would happen if no twist occurs.
Finally, note that $\kappa \in \mathbb{R}$ enforces other restrictions, see Grassmann \textit{et al.} \cite{GrassmannSenykBurgner-Kahrs_arXiv_2024}.

Second, \eqref{eq:rhoclarke_arc_parameters_constant_curvature} uses $\kappa = \textit{const.}$, whereas \eqref{eq:rhoclarke_arc_parameters} utilize the tip orientation called azimuth and elevation angles, \textit{i.e.}, $\theta$ and $\pi - \phi$, respectively.
Due to the relationship with the parallel curve, $\phi$ is related to $\rhovec$.
Therefore, it is unnecessary to consider $\phi = l\kappa$ as a relation to constant curvature. 

Third, \eqref{eq:rhoclarke_arc_parameters} avoids the parametrization using only $\theta$ and $\phi$ (or $\kappa$) by combining them in a non-linear way.
It has been shown that linear parametrization can demonstrate pathological behaviors, whereas a nonlinear parameterization, \textit{ that is,}, $\kappa\cos\left(\theta\right)$ and $\kappa\sin\left(\theta\right)$, is preferable.
We kindly refer to Della Santina~\textit{et al.} \cite{DellaSantinaBicchiRus_RAL_2020} and Dupont~\textit{et al.} \cite{DupontRucker_et_al_JPROC_2022} for further insights.

Therefore, we advocate using \eqref{eq:rhoclarke_arc_parameters} as a parameterization of the arc space.
As laid out above, it has desirable advantages over the other parameterization. 
Furthermore, it has a closed-form robot-depending mapping between the joint space $\jointspace$ and arc space without assuming constant curvature.

\subsection{Numerical Singularities in Constant Curvature Kinematics}

When using Clarke coordinates, including improved state parametrizations, the robot-dependent mappings do not inherently introduce coordinate singularities. 
The forward robot-dependent mapping $f_\text{dep}$ is given directly by \eqref{eq:rhoclarke_arc_parameters}.
Its inverse, the inverse robot-dependent mapping $f_\text{dep}^{-1}$, is similarly straightforward \eqref{eq:inverse}.
However, numerical singularities may still arise from the subsequent application of robot-independent mappings, especially when the constant curvature assumption is involved. 
These singularities occur explicitly at straight configurations (i.e., when curvature $\kappa = 0$), causing divisions by zero or numerically ill-conditioned computations.

To address these singularities, several approaches have been proposed in the literature:
\begin{itemize}
    \item Avoiding straight configuration entirely ($\kappa \not= 0$).
    \item Adding a small enough positive number $\epsilon$ to $\kappa \geq 0$,
    \item Saturating $\kappa \geq 0$ such that $\epsilon$ is the minimum,
    \item Branching to a linear approximation around $\kappa = 0$.
    \item Branching to an analytic limiting case when $\kappa = 0$.
    \item Choosing a different coordinate frame.
\end{itemize}

Among these strategies, selecting an appropriate coordinate frame represents the most elegant solution, thoroughly discussed by Della Santina~\textit{et al.}\cite{DellaSantinaBicchiRus_RAL_2020}.
While Clarke coordinates have no known singularities, numerical singularities persist in the forward kinematics and Jacobian computations of continuum robots modeled under constant curvature assumptions\cite{DellaSantinaBicchiRus_RAL_2020, DianGuo_et_al_Access_2022, AllenAlbert_et_al_RoboSoft_2020, GrassmannSenykBurgner-Kahrs_arXiv_2024}.
A division characterizes those terms by $\rhoclarke\transpose\rhoclarke$ that is equivalent to a division by $\kappa$, $\theta$, or $\rhovec\transpose\rhovec$, see \eqref{eq:rhoclarke_arc_parameters_constant_curvature} and \eqref{eq:rhoclarke_arc_parameters}.
To avoid numerical issues, Della Santina \textit{et al.} \cite{DellaSantinaBicchiRus_RAL_2020} branch to the analytic solution, Allen \textit{et al.} \cite{AllenAlbert_et_al_RoboSoft_2020} use linearization, and Grassmann \textit{et al.} \cite{GrassmannSenykBurgner-Kahrs_arXiv_2024} add a small number $\epsilon$.
Specifically, singularities emerge due to the normalization of displacement vectors \(\rhovec\), a step mandated by actuator-dependent joint parameterizations. 
Importantly, these numerical issues are inherent to the normalization step required by displacement-actuated joint values, not a limitation of Clarke coordinates or improved state parameterizations themselves. 

To avoid discontinuity due to branching, we suggest using an adaptive version of adding a small number.
For example,
\begin{align}
    \left(\dfrac{2}{n}\rhovec\transpose\rhovec\right)^{1/2}
    \longleftarrow
    \left(\dfrac{2}{n}\rhovec\transpose\rhovec\right)^{1/2} + \epsilon f\left(a + b\rhovec\transpose\rhovec\right)
    \label{eq:singularity_avoidance}
\end{align}
where $f\left(a + b\rhovec\transpose\rhovec\right)$ is a smooth decaying function, \textit{i.e.}, 
\begin{align}
    0 < \lim_{\rhovec\transpose\rhovec \rightarrow 0}f\left(a + b\rhovec\transpose\rhovec\right) &\leq 1
    \quad\text{and}\quad
    \nonumber\\
    \lim_{\rhovec\transpose\rhovec \rightarrow \inf}f\left(a + b\rhovec\transpose\rhovec\right) &= 0
    ,
    \nonumber
\end{align}
with $a$ and $b$ as parameters to enforce a desired behavior.
Suitable candidates are exponential decay functions or mirrored logistic (sigmoid) functions.
Note that the left-hand side of \eqref{eq:singularity_avoidance} is equivalent to the square root of $\rhoclarke\transpose\rhoclarke$ as shown in Grassmann \textit{et al.} \cite{GrassmannSenykBurgner-Kahrs_arXiv_2024}.
Practitioners will recognize the left-hand side of \eqref{eq:singularity_avoidance} in previous work \cite{DellaSantinaBicchiRus_RAL_2020, DianGuo_et_al_Access_2022, GrassmannSenykBurgner-Kahrs_arXiv_2024} by several authors.
Using the concept of virtual displacement \cite{FirdausVadali_AIR_2023, GrassmannSenykBurgner-Kahrs_arXiv_2024}, the geometric interpretation of \eqref{eq:singularity_avoidance} and $\epsilon$ are clear.

\section{Future Directions}
\label{sec:possibilities_opportunities}

Having established Clarke coordinates as a generalized and unified representation for continuum robots, we highlight promising avenues for future exploration. 
While several opportunities related to trajectory generation, modular frameworks, and control have been extensively discussed in previous works \cite{GrassmannBurgner-Kahrs_ICRA_EA_2024,GrassmannBurgner-Kahrs_arXiv_2024a,GrassmannBurgner-Kahrs_arXiv_2024b}, here we emphasize two novel theoretical opportunities: (i) the concept of flow and virtual displacement, and (ii) the mathematical nature of Clarke Coordinates.

\subsection{Flow and Effort}

Effort and flow are parametrizations in a bond graph \cite{GawthropBevan_CSM_2007} generalizing voltage, force, stress, \textit{etc.} and current, velocity, stress, strain rate, \textit{etc.}, respectively. 
Currently, the time integral of the flow, \textit{i.e.}, displacement, is considered.
In our case, the displacement is denoted by $\rhovec$.
Fortunately, as \eqref{eq:MP} and \eqref{eq:MP_inverse} are time-invariant, we can write
\begin{align}
    \dot{\rhoclarke} &= \MP\dot{\rhovec}
    \quad\text{and}\quad
    \label{eq:forward_dt}
    \\
    \dot{\rhovec} &= \MPinv\dot{\rhoclarke}
    ,
    \label{eq:inverse_dt}
\end{align}
where it combination reveals a velocity constraint, \textit{i.e.},
\begin{align}
    \dot{\rhovec} &= \MPinv\MP\dot{\rhovec}.
\end{align}
This has implications that will be the subject of future work.

Regarding the consideration of effort, we mention that \eqref{eq:M_Clarke} is also used for effort, \textit{i.e.}, voltage, in our previous discussion on electrical systems in Sec.~\ref{sec:intuition}.
Moreover, in the literature, we find several indications that the Clarke transform is the right transformation to consider the constraint and reduce the dimensionality.
Especially, work by Camarillo \textit{et al.} \cite{CamarilloSalisbury_et_al_TRO_2008, CamarilloCarlsonSalisbury_TRO_2009}, Dalvand \textit{et al.} \cite{DalvandNahavandiHowe_TRO_2018, DalvandNahavandiHowe_Access_2020, DalvandNahavandiHowe_Access_2022}, and Olso \textit{et al.} \cite{OlsonMenguc_et_al_IJSS_2020, OlsonMenguc_et_al_IJRR_2021, OlsonAdamsMenguc_BB_2022} indicates a constraint dual to the displacement constraint \eqref{eq:sum_rho}, where the actuation forces $\tau_i$ are constraint by
\begin{align}
	\sum_{i=1}^{n} \tau_{i} = 0.
    \nonumber
\end{align}
This gives rise to investigating the use of Clarke transform for $\tau_i$ related to effort.
Another hind is given by Allen \textit{et al.} \cite{AllenAlbert_et_al_RoboSoft_2020} mentioning that calculating torque in terms of their improved state parametrization is significantly more intuitive.

For example, Hsiao \& Mochiyama \cite{HsiaoMochiyama_IROS_2017} present a variant of a Clarke transformation matrix \eqref{eq:M_Clarke}, although they do not explicitly recognize it.
They relate strains due to wire-pulling to a bending via a constant matrix, which can be identified as \eqref{eq:M_Clarke} with $k_0 = 1$ and $k_1 = 0$ as well as row and column permutations.
By setting $k_1 = 0$, a torsionless segment is enforced.
Unfortunately, this choice prevents a straightforward inversion of their constant matrix, which is not provided.
Furthermore, this constant matrix \cite{HsiaoMochiyama_IROS_2017} is neither amplitude-invariant nor power-invariant as indicated by choice $k_0 = 1$.
Future work will shed more light on the application of these efforts.

\subsection{The Mathematical Nature of Clarke Coordinates}

The Clarke transform relates to the field of electric systems as the analogy in Grassmann \textit{et al.} \cite{GrassmannSenykBurgner-Kahrs_arXiv_2024} suggests.
This field is rich in mathematical descriptions, \textit{e.g.}, phasors, and complex numbers, that can be transferred and applied to displacement-actuated continuum robots.
In particular, the following identities of the encoder-decoder architecture \cite{GrassmannBurgner-Kahrs_arXiv_2024a} should be more explored.
While easy to construct from \cite{GrassmannSenykBurgner-Kahrs_arXiv_2024, GrassmannBurgner-Kahrs_ICRA_EA_2024}, but not stated before, each of them relates to transformation of different inputs, \textit{i.e.}, $\rhovec$, $\hotvec_{n \times 1}^{(k)}$, or $\onevec_{n \times 1}$, and the Toeplitz matrix $\MPinv\MP \not = \Imat_{n \times n}$.

Used in our definition of $\jointspace$ in \eqref{eq:jointspace_alt_wo_ClarkeCoordinates}, the first identity is
\begin{align}
    \rhovec &= \MPinv\MP\rhovec
    \nonumber
\end{align}
that is a reminder of a transformation of eigenvectors with an eigenvalue equal to one.
Applying the above equation in a sequence shows that $\MPinv\MP\rhovec$ is an idempotent matrix.
It should be noted that $\rhovec \in \jointspace$ is mapped to the same $\rhovec$.

The second identity is given by
\begin{align}
    \begin{bmatrix}
        \cos\left(\psi_1 - \psi_k\right)\\
        \cos\left(\psi_2 - \psi_k\right)\\
        \vdots \\
        \cos\left(\psi_n - \psi_k\right)
    \end{bmatrix}
    &= \MPinv\MP\hotvec_{n \times 1}^{(k)}
    ,
    \nonumber
\end{align}
where $\hotvec_{n \times 1}^{(k)}$ is a one-hot vector defined by the $k\textsuperscript{th}$ element to be one, whereas all other elements are zero.
Note that the input gets amplified. 
Using the trigonometric identities \cite{GrassmannSenykBurgner-Kahrs_arXiv_2024}, the magnitude of left-hand side is $\sqrt{n/2} > 1$ for $n \geq 3$, whereas the magnitude $\hotvec_{n \times 1}^{(k)}$ is one.

The last identity is 
\begin{align}
    \mathbf{0}_{n \times 1} &= \MPinv\MP\onevec_{n \times 1}
    ,
    \nonumber
\end{align}
where $\onevec_{n \times 1}$ has ones everywhere and $\mathbf{0}_{n \times 1}$ has zeros everywhere.
Note that the input is filtered, \textit{i.e.}, $\onevec_{n \times 1}$ is mapped to $\mathbf{0}_{n \times 1}$.

Those identities are important to understand error propagation in the case where the displacement is not an element of the joint space $\jointspace$.
They might be useful in an optimization framework as well.
Future work should explore, apply, and transfer those and similar mathematical descriptions.

\section{Conclusion}

We have demonstrated that previously proposed improved state parameterizations for various continuum and soft robots are special cases of Clarke coordinates. 
By unifying these representations through a generalized Clarke transformation matrix, we consolidate multiple benefits into a single mathematical framework. 
This generalization explicitly incorporates arbitrary joint arrangements, numbers of joints, and underlying modeling assumptions, thereby significantly broadening applicability and enabling novel robot designs and analyses. 
Our unified approach thus provides a robust theoretical foundation for advancing continuum and soft robotics research.









\addcontentsline{toc}{section}{REFERENCES}
\bibliographystyle{IEEEtran}
\bibliography{IEEEabrv, literature}


\end{document}